\documentclass{article}

\usepackage{microtype}
\usepackage{graphicx}
\usepackage{subcaption}
\usepackage{booktabs}
\usepackage{amsmath,amssymb,amsfonts,amsthm}
\usepackage{bm}
\usepackage{xcolor}
\usepackage{algorithm}
\usepackage[noend]{algpseudocode} 
\usepackage{authblk}
\usepackage{cleveref}

\theoremstyle{plain}
\newtheorem{theorem}{Theorem}[section]

\newtheorem{corollary}[theorem]{Corollary}
\newtheorem{lemma}[theorem]{Lemma}
\newtheorem{assumption}[theorem]{Assumption}

\theoremstyle{definition}
\newtheorem{definition}[theorem]{Definition}

\theoremstyle{remark}
\newtheorem*{remark}{Remark}

\DeclareMathOperator*{\argmin}{arg min} 
\DeclareMathOperator*{\argmax}{arg max} 

\newcommand{\vct}[1]{\bm{#1}}
\newcommand{\mtx}[1]{\bm{#1}}
\newcommand{\norm}[1]{\lVert {#1} \rVert}

\def \Re {\mathbb{R}}

\def \X {\mtx{X}}
\def \1 {\vct{1}}
\def \x {\vct{x}}

\def \z {\vct{z}}
\def \w {\vct{w}}

\def \ws {\vct{w}_{\lambda_{s}}^*}
\def \wl {\vct{w}_{\lambda}^*}
\def \barws {\overline{\vct{w}}_s}
\def \barw {\overline{\vct{w}}}

\def \Fls {F_{\lambda_s}}

\def \Fl {F_\lambda}

\def\acks#1{\vskip 0.3in\noindent{\large\bf Acknowledgments}\vskip 0.2in
\noindent #1}

\Crefname{subsection}{Subsection}{Subsections}
\Crefname{assumption}{Assumption}{Assumption}

\begin{document}

\title{Bias of Homotopic Gradient Descent for the Hinge Loss}

\author[1]{Denali Molitor}
\author[1]{Deanna Needell}
\author[2]{Rachel Ward}
\affil[1]{Department of Mathematics, University of California at Los Angeles, Los Angeles, California, USA}
\affil[2]{Department of Mathematics, University of Texas at Austin, Austin, Texas, USA}

\maketitle

\vskip 0.3in
\begin{abstract}
Gradient descent is a simple and widely used optimization method for machine learning. For homogeneous linear classifiers applied to separable data, gradient descent has been shown to converge to the maximal margin (or equivalently, the minimal norm) solution for various smooth loss functions.  The previous theory does not, however, apply to non-smooth functions such as the hinge loss which is widely used in practice. Here, we study the convergence of a homotopic variant of gradient descent applied to the hinge loss and provide explicit convergence rates to the max-margin solution for linearly separable data. 
\end{abstract}

\section{Introduction}

Several recent works suggest that the optimization methods used in training models affect the model\rq{}s ability to generalize through implicit biases to certain solutions \cite{Zhang2017,neyshabur2014search,Hardt2016,hoffer2017train,2018arXiv180100173P,2018arXiv180611379P,hoffer2017train,Chaudhari2017,combes2018learning}. 
 In order to understand the effects of optimization methods in more complex and often non-convex settings such as for neural networks, it is natural to first understand their behavior in simpler settings, such as for least squares regression, logistic regression, and support vector machines (SVM) \cite{soudry2018implicit,nacson2018convergence,gunasekar2018characterizing}. In particular, gradient descent and its many variants, including the subgradient method, are popular choices for optimizing machine learning models and thus warrant careful study. 

It was recently shown that gradient descent applied to the (unregularized) logistic regression problem for linearly separable data converges to the solution with maximal margin, while other choices of optimization method converge to different solutions \cite{soudry2018implicit}. Convergence to the maximal-margin solution is desirable, as the margin is an important quantity for deriving generalization guarantees \cite{bartlett1998generalization,Vapnik1982,vapnik1999overview,vapnik1974theory,vapnik2013nature}. The analysis of Soudry et al \cite{soudry2018implicit} extends to additional loss functions, but requires particular properties, including  smoothness and monotonicity. These assumptions do not hold, however, for non-differentiable functions such as the hinge loss objective, which is the loss function used in training SVM \cite{cortes1995support}. 

Here, we analyze the convergence to the maximal margin solution of a homotopic subgradient method applied to the non-smooth hinge loss. In particular we consider a method in which a number of subgradient updates are applied to the hinge loss with decreasing regularization. Although it is well known that the exact solutions of the regularized hinge loss converge to the hard-margin SVM solution as the regularization decreases to zero in the linearly separable case \cite{rosset2004margin,hastie2004entire}, we are unaware of results that provide explicit convergence rates for an iterative optimization algorithm, such as the subgradient method, that converges to the hard-margin SVM solution in a single pass of the regularization parameter $\lambda$. We provide such an analysis here, and demonstrate that the iterates of an averaged subgradient method applied to the regularized SVM loss with shrinking regularization parameters converge to the max-margin solution at a rate of $O\left(k^{-1/6+\delta}\right)$ for linearly separable data, where $\delta$ is any small positive constant. 

For linearly separable data there exists $\lambda' >0$ be such that the solution $\wl$ to the hinge loss with regularization parameter $\lambda$ is equal to the true,  hard-margin solution $\w^*$ for all $\lambda \le \lambda'$ \cite{rosset2004margin,hastie2004entire}. While $\lambda\rq{}$ is constant for a fixed problem, knowing its value in advance is typically unrealistic. Additionally, if the data is not well separated, $\lambda\rq{}$ can be very small. The homotopic subgradient method analyzed here depends on the value of $\lambda\rq{}$ and converges at a rate of $O\left((\lambda\rq{})^{-2} k^{-1/6+\delta}\right)$.
If one were to know the appropriate regularization parameter $\lambda\rq{}$ in advance, the averaged subgradient method with appropriate fixed step sizes would converge in $L_2$ error at a rate of $O\left((\lambda')^{-1}k^{-1/4}\right)$. This rate can be improved to $O\left((\lambda')^{-1}k^{-1/2}\right)$ by using weighted step sizes that depend on $\lambda\rq{}$ \cite{bubeck2015convex,lacoste2012simpler}. Thus, we pay a small price for the shrinking regularization routine and for not knowing the value of $\lambda\rq{}$ in advance. We additionally provide faster convergence guarantees and improved convergence results for the proposed method on small datasets as compared to gradient descent applied to the logistic loss with fixed step sizes \cite{soudry2018implicit}.

\subsection{Contributions} 
While several works analyze the convergence of various optimization methods to the maximal-margin solution for separable data \cite{soudry2018implicit,nacson2018convergence}, we are unaware of any works that provide explicit convergence rates for the fundamental subgradient descent method. Convergence of the subgradient method and stochastic subgradient method have been analyzed for non-smooth convex functions, however these works only provide convergence guarantees in the loss-function values and not the iterates, as, for general convex functions, the minimizer may not be finite and may not be unique \cite{shamir2013stochastic,zhang2004solving}. In the context of solving the hard-margin SVM, the restriction to linearly separable data guarantees the existence of a minimizer and considering the maximal margin solution ensures uniqueness. Moreover, in the context of general convex functions, previous works often use the projected subgradient method and require knowledge of a bounded domain in which a minimizer exists \cite{shamir2013stochastic, bubeck2015convex}. For solving the hard-margin SVM via gradient descent, we show that such a projection is unnecessary.

Here, we provide explicit convergence guarantees for a homotopic subgradient method for optimizing the non-smooth SVM hinge loss. The proposed method uses decreasing regularization parameters and leads to the hard-margin SVM solution. We study the effects of optimization via this method on the generalization ability of the learned solutions through proved convergence rates to the hard-margin SVM solution in terms of $L_2$ error as well as difference in angle and margin from the true solution. We additionally show that these convergence rates to the hard-margin SVM solution outpace recent results such as gradient descent with fixed step sizes applied to the logistic loss \cite{soudry2018implicit,nacson2018convergence}. We demonstrate the convergence of the proposed method on a synthetic dataset.

\subsection{Organization}
In \Cref{sec:setup}, we introduce the specific problem setting, the notation that will be used throughout, and the proposed optimization scheme, \Cref{algo:GD}. \Cref{sec:Main} provides the main convergence results for \Cref{algo:GD}. An outline for the proof of the main convergence theorem, \Cref{thm:conv}, is provided in \Cref{sec:MainProof}, with additional details in \Cref{sec:lemma_proofs}.  We test convergence properties of \Cref{algo:GD} for a simple synthetic dataset in \Cref{sec:exper}. \Cref{sec:implement} provides additional implementation details for \Cref{algo:GD} as well as possible modifications and extensions.

\section{Problem Setup}
\label{sec:setup}

We consider the binary classification problem with data $\{(\x_j, y_j) : j=1,\ldots,n\}$, where $\x_j\in\Re^d$ are the data points and $y_j\in\{-1, 1\}$ their labels. We aim to classify the data via a homogeneous linear SVM. Specifically, we wish to identify a weight vector $\w^*$ that satisfies 
\[
\w^* = \argmin_{\w} \|\w\| \quad \text{ subject to } \quad y_j \x_j^\top  \w \ge 1 \; \forall \, j.
\]
Throughout, we write $\|\cdot\|=\|\cdot\|_2$. 
We can equivalently find
\begin{equation}
\w^* = \argmin_{\w} \|\w\|  \text{ subject to } \mathcal{L}(\w) = 0,
\label{eqn:hinge0}
\end{equation}
where 
\[ 
\mathcal{L}(\w) :=\frac{1}{n}\sum_{j=1}^n h(y_j\x_j^\top \w)  \text{ and }  h(u) := \max(0, 1-u).
\]
The function $h(u)$ is commonly referred to as the \emph{hinge loss}. We assume throughout that the data is linearly separable, i.e. there exists a vector $\w$ satisfying $\mathcal{L}(\w) = 0$ as is done in \cite{soudry2018implicit,nacson2018convergence,Wang2019ReLULinSep,iclr2019Brutzkus,pmlr-v89-nacson19a,rosset2004margin}. This assumption is common and necessary in order to discuss the margin of the approximated solutions. Minimizing the norm of the solution $\w$ to $\mathcal{L}(\w)=0$ corresponds to maximizing the margin, that is maximizing the minimal distance between any data point and the separating hyperplane determined by $\w$. In this setting, the solution to \Cref{eqn:hinge0}, $\w^*$, is often referred to as the {\it hard-margin SVM solution}.

The constrained optimization problem in \Cref{eqn:hinge0} is the primal formulation of an SVM. While solving or approximating the corresponding dual SVM formulation is popular in practice, there are advantages to approximating the primal problem directly \cite{Chapelle2007Primal}. Of particular interest for this work, considering the primal formulation allows for straightforward analysis of the effect of the optimization error on the margin and hyperplane angle.

As an alternative to solving \Cref{eqn:hinge0} directly, one often looks for a solution to an unconstrained, regularized version. Define the functional:
\begin{equation}
\label{eqn:LShinge}
F_{\lambda}(\w) := \frac{\lambda}{2} \| \w \|^2 + \frac{1}{n} \sum_{j=1}^n h(y_j\x_j^\top \w) .
\end{equation} 
For $\lambda > 0$, $\Fl$ is strongly convex with strong convexity parameter $\lambda$. We will use $\partial F$ to denote the subgradient of $F$. The gradient of $F_{\lambda}(\w)$ exists as long as $y_j\x_j^\top \w \neq 1$ for all $j$ and is given by
\begin{equation}\label{eqn:subgrad}
 \partial F_{\lambda}(\w) = \lambda \w - \frac{1}{n} \sum_{j: y_j \x_j^\top \w < 1} y_j \x_j.
\end{equation}
When $y_j\x_j^\top \w = 1$ for some $j$, the subgradient set $\partial F_{\lambda}(\w)$ contains the point 
\[
\lambda \w - \frac{1}{n} \sum_{j: y_j \x_j^\top \w < 1} y_j \x_j \in \partial F_{\lambda}(\w) .
\]
When the gradient does not exist, we will abuse notation and use \Cref{eqn:subgrad} in the subgradient method update of \Cref{eqn:subgradMethod}. 

Let 
\begin{equation}\label{eqn:subprob}
\wl := \argmin_{\w} \Fl(\w).
\end{equation}
We will refer to $\wl$ as the solution to the regularized subproblem of minimizing \Cref{eqn:LShinge}. A larger regularization parameter $\lambda$ encourages a solution $\wl$ with smaller norm at the cost of having some points lie within the margin. For linearly separable data and as $\lambda$ approaches 0, the regularized solutions $\wl$ converge to the unregularized solution, $\w^*$. Let $\lambda' >0$ be such that $\wl = \w^*$ for all $\lambda \le \lambda'$. Such a $\lambda'$ is guaranteed to exist for linearly separable data \cite{rosset2004margin,hastie2004entire}. This fact suggests solving \Cref{eqn:LShinge} by using the subgradient method for a sufficiently small value of $\lambda$. Of course, the value of $\lambda'$ will typically be unknown. 

We use the following assumption and definition of $\lambda\rq{}$ throughout.
\begin{assumption}
The data $\x_1, \ldots, \x_n\in \mathbb{R}^d$ with labels $y_1,\ldots, y_n\in\{-1,1\}$ are linearly separable, i.e.\ there exists $\w$ such that for all $i$, $y_i \w^\top  \x_i >0$. Let $\w^*$ be the hard-margin SVM (i.e. $\w^*$ solves \Cref{eqn:hinge0}) and $\lambda'$ be such that for all $\lambda \le \lambda'$, $\wl=
\argmin \Fl = \w^*$.
\label{assumption:linSep}
\end{assumption}

While in practice, one may be satisfied with the solution $\wl$ for $\lambda$ sufficiently small, we are interested in the convergence to the true hard-margin SVM given by $\w^*$. Thus, we instead propose to use a ``homotopic" variant of the subgradient method that iteratively approximates the solution to  \Cref{eqn:LShinge} while the regularization parameter $\lambda$ and accompanying step size $\eta$ of the subgradient method in \Cref{eqn:subgradMethod} decay at prescribed rates. Incorporating a piecewise constant decaying step size is commonly used for large-scale minimization problems, especially when using stochastic gradient descent variants \cite{bottou2018optimization}.

Recall the subgradient method given by the updates:
\begin{equation}\label{eqn:subgradMethod}
\w_{k+1} = \w_k - \eta_k\partial\Fl(\w_k),
\end{equation}
where $\w_k$ is the approximate solution at iteration $k$ and $\eta_k$ is a step size. 
For some number of outer iterations $s = 1, \dots, S$, we choose a regularization parameter $\lambda_s > 0$, a step size $\eta_s > 0$ and a number of inner iterations $t_s$. The regularization parameter $\lambda_s$ and step size $\eta_s$ are selected such that they decrease to 0 as $s$ increases. Let $\barw_{s-1}$ be the current estimate of $\w^*$. We then perform $t_s$ subgradient updates applied to the loss function $\Fls$ with initial iterate $\barw_{s-1}$ and step size $\eta_s$. The next estimate, $\barws$, is given by the average of the $t_s$ subgradient iterates. This process is detailed in \Cref{algo:GD}. For specific choices of $\lambda_s,$ $\eta_s$ and $t_s$, \Cref{algo:GD} converges to the hard-margin SVM solution $\w^*$. Convergence guarantees are detailed in \Cref{thm:conv}.

\begin{algorithm}
	\caption{Homotopic subgradient method }
	\label{algo:GD}
\begin{algorithmic}
\State {\bfseries Input:} data $\{\x_j\}$, labels $\{y_j\}$, maximum outer iterations $S$, parameter for initial inner iterations $s_0>2$, parameters $1>p>0$ and $r>2p$
\State{Define $\epsilon_0 = \frac{ \log(s_0 ) - \log(s_0 -1)}{\log(s_0)}$,  $\alpha = \min\left\{\frac{r-2p}{2(1+\epsilon_0)},1-p\right\}$, and $C = \max\left\{4, \tfrac{1}{2 }s_0^p(s_0-1)^{\alpha}\right\}$}
\State Initialize $\barw_{0}= \bf{0}$
\For {$s=0,1,\ldots, S-1  $ }
	\State  $\lambda_s = (s_0+s)^{-p}$, $t_s = (s_0+s)^{r}$,   and $\eta_s = \frac{C(s_0+s-1)^{-\alpha}}{\sqrt{t_s}}$
	\State   $\w_0 = \barws$
	\For {$i = 1, 2, \ldots, t_s$}
		\State  $\w_{i} = (1-\lambda_s\eta_s) \w_{i-1} + \frac{\eta_s}{n}\sum_{j : y_j\x_j^\top  \w_{i-1}\leq 1} y_j\x_j$ \label{eqn:iters}
	\EndFor
	\State  $\barw_{s+1} = \frac{1}{t_s} \sum_{i=1}^{t_s} \w_{i}$
\EndFor
\State Output $\barw_S$
\end{algorithmic}
\end{algorithm}

While the strongly convex functions $\Fl$ are not globally Lipschitz, they are Lipschitz functions on bounded domains. Using a projected subgradient method in which iterates are projected onto a bounded domain is a natural strategy for restricting the domain of the iterates. A projection is unnecessary in this setting, however, as the regularization parameter $\lambda > 0$ naturally promotes solutions of smaller norm. 
In fact, the iterates produced by the subgradient method in \Cref{algo:GD} remain bounded in norm with a bound that depends on the current regularization parameter $\lambda$. 

\begin{lemma}\label{lem:iter_bound}
Fix a regularization parameter $\lambda > 0$ and step size $\eta > 0$ such that $\eta\lambda < 1$. Define 
\begin{equation}\label{eqn:iter_bound}
 B_\lambda := \frac{\sum_{j=1}^n\norm{\x_j }}{\lambda n}.
\end{equation}
If the initial iterate $\w_0$ is such that $\norm{\w_0} \le B_\lambda$, then each iterate $\w_k$ produced by the subgradient method of \Cref{eqn:subgrad} applied to the function $\Fl$ of \Cref{eqn:LShinge} has $\norm{\w_k} \le B_\lambda$. Additionally, $\norm{\w^*}\le B_\lambda$.
\end{lemma}
 In summary, if the initial iterate $\w_0$ is such that $\norm{\w_0}\le B_\lambda$, then the iterates produced by the subgradient method applied to $\Fl$ will also have norm less than or equal to $B_\lambda$.

\begin{remark}
Using \Cref{lem:iter_bound}, one can show that the functionals $\Fl$ are Lipschitz over the domain of iterates produced by \Cref{algo:GD}. Specifically, the constant 
\begin{equation}\label{eqn:L}
L :=  \frac{2}{n} \sum_{j=1}^n{\| \x_j \|} 
\end{equation}
bounds the Lipschitz constants of each function $\Fl$ restricted to the ball centered at the origin with radius $B_\lambda$. \Cref{lem:iter_bound} guarantees that the iterates produced when applying the subgradient method to $\Fl$ and for sufficiently small initial iterate remain with this domain. Note that the bound on the Lipschitz constants $L$ is independent of the regularization parameter $\lambda$.
\end{remark}

\section{Main Results}
\label{sec:Main}

We now provide explicit rates of convergence to the hard-margin SVM solution for \Cref{algo:GD}. We provide convergence rates in terms of the $L_2$ error, difference in angle, and difference in margin between the approximation $\barw_S$ and the true hard-margin solution. 
The convergence results are stated in terms of $k$, the total number of subgradient updates required. Recall that the approximations $\barws$ are only updated at increments of $t_s$ subgradient updates. Let $\z_k = \barws$, so that $\z_k$ is the approximation after $k = \sum_{i=1}^s t_i$ subgradient calculations.

\Cref{thm:conv} provides a convergence guarantee for the $L_2$ error of the iterates produced by \Cref{algo:GD}. This result will be used to additionally derive convergence guarantees for the angle and margin of the solution in \Cref{lem:angleMargin}. The parameter $p$ determines the rate of decay of the regularization $\lambda_s$ and the parameter $r$ determines the number of steps $t_s$ used at each fixed level of regularization. The constant $L$ is as defined in \Cref{eqn:L} and is an upper bound on the Lipschitz constants of the functions $\Fl$ restricted to the domain of the iterates produced by the subgradient method applied to $\Fl$ (\Cref{lem:iter_bound}).
 
\begin{theorem}
\label{thm:conv} 
Consider \Cref{algo:GD} with parameters $r$ and $p$ such that $0<p<1$ and $r>2p$. Choose an initial number of inner iterations $s_0^r \in \mathbb{N}$ with $s_0>2$. 
Let $L = 2\frac{\sum_{j=1}^n \norm{\x_j}}{n}$ as defined in \Cref{eqn:L}. 
Define
\[
C = \max\left\{4, \tfrac{1}{2 }s_0^p(s_0-1)^{\alpha}\right\}\quad\text{and}\quad \alpha = \min \left(\frac{r-2p}{2(1+\epsilon_0)}, 1 - p  \right) ,\]
with 
$
\epsilon_0 = \frac{ \log(s_0 ) - \log(s_0 -1)}{\log(s_0)}.
$
Let $\z_k$ be the average of the $t_s$ subgradient descent updates calculated to minimize the function $\Fls$ with step size $\eta_s = \frac{C(s_0+s-1)^{-\alpha}}{\sqrt{t_s}}$, where $k$ is the total number of subgradient descent updates calculated. 
Then for data and $\lambda\rq{}$ satisfying \Cref{assumption:linSep}, 
\begin{equation}\label{eqn:conv32_2terms}
\|\z_k - \w^*\| 
\le CL \left((r+1)k\right)^{\frac{-\alpha(1-\epsilon_0)}{r+1}} 
+  \frac{L  }{2(\lambda\rq{})^2} \left((r+1)k\right)^{\frac{-p}{r+1}}.
\end{equation}
Let $c = \min\left(\frac{\alpha(1-\epsilon_0)}{r+1}, \frac{p}{r+1}\right)$. Then 
\begin{align*}
\|\z_k - \w^*\| 
&\le \left(C + \frac{1  }{2(\lambda\rq{})^2} \right) L (r+1)^{-c}k^{-c}.
\end{align*}
\end{theorem}

An outline for a proof of \Cref{thm:conv} can be found in \Cref{sec:MainProof} with additional details in \Cref{sec:lemma_proofs}. Note that, for small $\epsilon_0$, the two terms in the bound of \Cref{eqn:conv32_2terms} will decrease at approximately the same rate if $r=2$ and $p=1/2$. 
\Cref{cor:expl_conv} gives a simpler, explicit rate of convergence by making this specification and setting $s_0 = 10$.
\begin{corollary} 
\label{cor:expl_conv}
Consider \Cref{algo:GD} with parameters $r=2$, $p=1/2$ and an initial number of inner iterations $s_0^r = s_0^2 \in \mathbb{N}$ with $s_0>2$. 
Let $L = 2 \frac{\sum_{j=1}^n \norm{\x_j}}{n}$.
Define
\[
C = \max\left\{4, \tfrac{1}{2 }s_0^p(s_0-1)^{\alpha}\right\}\quad\text{and}\quad \alpha = \min \left(\frac{r-2p}{2(1+\epsilon_0)}, 1 - p  \right) ,\]
with 
$
\epsilon_0 = \frac{ \log(s_0 ) - \log(s_0 -1)}{\log(s_0)}.
$
Let $\z_k$ be the average of the $t_s$ subgradient descent updates calculated for $\Fls$ with step size $\eta_s = \frac{C(s_0+s-1)^{-\alpha}}{\sqrt{t_s}}$, where $k$ is the total number of subgradient descent updates calculated. 
Then for data and $\lambda\rq{}$ satisfying \Cref{assumption:linSep}, 
\begin{align*}
\|\z_k - \w^*\| 
&\le CL \left(3k\right)^{\frac{-1}{6}\frac{(1-\epsilon_0)}{(1+\epsilon_0)}} 
+  \frac{L \left(3k\right)^{-1/6} }{2(\lambda\rq{})^2}.
\end{align*}
\end{corollary}
Choosing $s_0=10$, we have $\epsilon_0 < 0.046$, $C < 4.9$ and arrive at the convergence rate
\[
\|\z_k - \w^*\|
 \le   4.17 L k^{\frac{-0.913}{6}} 
+  \frac{0.42L k^{-1/6} }{(\lambda\rq{})^2}.
\]
At least theoretically, sending $s_0\to\infty$ leads to the best convergence rate guarantee. In fact, the convergence rate provided by \Cref{thm:conv} can be made arbitrarily close to $O\left(k^{-1/6}\right)$ by choosing $r=2$, $p=1/2$, and $s_0$ sufficiently large. As we will see in \Cref{sec:exper}, using $s_0$ extremely large becomes impractical as the number of iterations for each fixed-$\lambda$ subproblem becomes extremely large. 

For strongly-convex, Lipschitz functions with strong-convexity parameter $\lambda$, one can achieve convergence in $\|\w-\w^*\|$ at a rate of $O\left(\lambda^{-1} k^{-1/4}\right)$, using projected averaged gradient descent with fixed step sizes (Theorem 3.2 \cite{bubeck2015convexarxiv}). Using weighted step sizes, and knowledge of the strong convexity parameter, this rate can be improved to $O\left(\lambda^{-1} k^{-1/2}\right)$ (Theorem 3.9 \cite{bubeck2015convexarxiv}, originally from \cite{lacoste2012simpler}). A challenge of solving for the hard-margin SVM is that we do not optimize a strongly convex function. While one could fix a regularization parameter $\lambda$ leading to a strongly convex function, there is no guarantee that the minimizer of this function $\Fl$ will correspond to the true solution $\w^*$. Since the convergence rate of \Cref{algo:GD} can be made arbitrarily close to $O\left( (\lambda')^{-2}k^{-1/6}\right)$ we lose very little, only a factor of $O\left( (\lambda')^{-1}k^{1/12+\delta}\right)$ compared to the convergence rate of projected averaged gradient descent with fixed step sizes, for not knowing $\lambda'$ in advance and instead incorporating decreasing explicit regularization. 

Additionally, in designing \Cref{algo:GD}, we aimed for a simple algorithm as opposed to optimizing all possible parameters. One could possibly improve on the rates given here by further optimizing these parameters.

\subsection{Convergence rates for angle and margin gaps}
\label{subsec:gap_conv}
The convergence rate in \Cref{thm:conv} can be used to derive rates of convergence to the angle and margin of the optimal separating hyperplane $\w^*$.  
\begin{definition}
For the hard margin SVM solution $\w^*$ and a vector $\w$, define 
\begin{equation}
\text{\emph{angle gap}} :=  1 - \frac{\w^\top \w^*}{\|\w\| \|\w^*\|}
\label{eqn:angle}
\end{equation}
and 
\begin{equation}
\text{\emph{margin gap}} := \frac{1}{\|\w^*\|} - \min_i \frac{y_i\x_i^\top  \w}{\|\w\|}.
\label{eqn:margin}
\end{equation}
\label{defn:gap}
\end{definition}
While it is natural to consider the $L_2$ error of the derived solution, the angle between the true and derived solutions as well as the difference in the size of the margins give a more intuitive interpretation of the effect of that error. For example, an approximate solution $\w$ that is off by a constant factor, that is $\w = c\w^*$, will have an angle gap of zero and non-zero margin gap if $c\ne 1$. If an approximate solution $\w$ has a nonzero angle gap, but negligible margin gap, this suggests that the derived solution $\w$ still separates the data reasonably well. 

Convergence rates of \Cref{algo:GD} in terms of the angle and margin gaps are stated in \Cref{lem:angleMargin} and compared to other recently obtained convergence rates in \Cref{tab:rates}. The rates of convergence in these metrics can be derived from \Cref{thm:conv}. These arguments are included in \Cref{sec:lemma_proofs}.

\begin{lemma}
\label{lem:angleMargin}
Let 
\[c =  \min \left(\frac{(r-2p)(1-\epsilon_0)}{2(r+1)(1+\epsilon_0)}, \frac{(1 - p)(1+\epsilon_0)}{r+1} , \frac{p}{r+1}  \right), \]
where $p, r, s_0$, and $\epsilon_0$ are as given in \Cref{thm:conv}
 so that $c$ is the exponent in the convergence rate of \Cref{thm:conv}. Let $\delta$ be such that $c = 1/6 - \delta$. The value of $\delta$ is positive and can be made arbitrarily close to 0 by choosing $s_0$ sufficiently large and setting $p=1/2$ and $r = 2.$ Then for the angle gap,
\[1 - \frac{\w_k^\top \w^*}{\|\w_k\| \|\w^*\|} = O\left(k^{-1/3 +2\delta}\right).\]
For the margin gap,
\[
\frac{1}{\|\w^*\|}  - \min_i \frac{y_i \x_i^\top  \w_k}{\|\w_k\|}  = O\left(k^{-1/6+\delta}\right). 
\] 
\end{lemma}

The convergence guarantees for the angle and margin gaps for \Cref{algo:GD} are significantly faster than 
those given in Soudry et al \cite{soudry2018implicit} for gradient descent with fixed step sizes applied to the logistic loss (see \Cref{tab:rates}). 
Nacson et al \cite{nacson2018convergence} demonstrate that using aggressive adaptive step sizes for gradient descent applied to the logistic loss  leads to 
a faster convergence rate of $O\left(\frac{\log(t)}{\sqrt{t}}\right)$. While the convergence guarantees for \Cref{algo:GD} are slower, as $c\le 1/6$, in this paper, we are interested in analyzing convergence guarantees for gradient descent applied to the non-smooth hinge loss.

\begin{table}
\begin{center}
\def\arraystretch{1.2}

\resizebox{0.6\linewidth}{!}{
\begin{tabular}{ |c|c|c| } 
 \hline
  &  \Cref{algo:GD} &  
 \cite{soudry2018implicit}\\
\hline \hline
 Angle gap 
& $O\left(k^{-1/3 + 2\delta}\right)$ 
& $O\left(\left(\frac{\log \log(k)}{\log(k)}\right)^2 \right) $ 
\\
\hline
 Margin gap  
& $O\left(k^{-1/6 +\delta}\right)$ 
& $O\left(\frac{1}{\log(k)}\right)$ 
\\
 \hline
\end{tabular}
}
\end{center}
\caption{Comparison of convergence rates for \Cref{algo:GD} with those of \cite{soudry2018implicit} for gradient descent with fixed step sizes applied to the logistic loss.}
\label{tab:rates}
\end{table}

\section{Proof of \Cref{thm:conv}}
\label{sec:MainProof}

We prove \Cref{thm:conv} through a series of lemmas, which are stated in \Cref{subsec:lemmas} and whose proofs are contained in \Cref{sec:lemma_proofs}. The proof of \Cref{thm:conv} is contained in \Cref{subsec:main_proof}.

We briefly summarize each of the lemmas for convenience. 
\Cref{lem:avg_bound} provides a modified convergence guarantee for the averaged subgradient method applied to the functions $\Fl$. \Cref{lem:gamma_bound} bounds the distance between minimizers of $\Fl$ for different regularization parameters $\lambda$. This result allows for the incorporation of the decreasing regularization in \Cref{algo:GD}. \Cref{lem:R_bound} makes use of \Cref{lem:avg_bound} and \Cref{lem:gamma_bound} to bound the initial error $\norm{\barws - \wl}$ of each regularized subproblem as given in \Cref{eqn:subprob}.

\subsection{Useful lemmas}\label{subsec:lemmas}

\Cref{lem:avg_bound} is a modified version of a standard convergence analysis of the averaged subgradient method for convex Lipschitz functions (Theorem 3.2 of \cite{bubeck2015convex}). This result bounds the distance between the average of the subgradient descent iterates $\barw$ and the minimizer $\wl$ of the functional $\Fl$ for a fixed regularization parameter $\lambda$. 

\begin{lemma}
\label{lem:avg_bound}
Let 
\[
\Fl(\w) = \frac{\lambda}{2} \| \w \|^2 + \frac{1}{n} \sum_{j=1}^n \max(0,1-y_j\x_j^\top \w)
\] 
and $L = 2 \frac{\sum_{j=1}^n \norm{\x_j}}{n}$.
 Let the initial iterate $\w_0$ be such that $\norm{\w_0} \le \frac{L}{2\lambda }$ and let $\wl$ minimize $\Fl$. Suppose $\|\w_0 - \wl \| \le R$, so that $\wl$ is contained in a ball of radius $R$ and center $\w_0$. Let $\barw = \frac{1}{t} \sum_{s=1}^t \w_s$ be the average of $t$ subgradient method iterates with initial iterate $\w_0$ and step size $\eta = \frac{R}{L\sqrt{t}}$. Then 
\[0\le F_{\lambda} (\barw) - F_{\lambda} (\w_{\lambda}^*) \le \frac{RL }{\sqrt{t}} - \frac{\lambda}{2}\| \barw - \wl\|^2.\]
\end{lemma}

Note that \Cref{lem:avg_bound} also guarantees that 
\[\| \barw - \wl\|^2 \le \frac{2RL }{\lambda\sqrt{t}} .\]

The next lemma bounds the distance between the minimizers $\wl$ and $\w_{\widetilde \lambda}^*$ of the functions $\Fl$ and $F_{\widetilde \lambda}$ and shows that distance from $\wl$ to the true hard-margin solution $\w^*$, $\|\wl - \w^*\|$, is proportional to the regularization parameter $\lambda$.
\begin{lemma}
\label{lem:gamma_bound}
Let $\wl$ minimize $\Fl$ as given in \Cref{eqn:LShinge} and let $\w^*$ solve \Cref{eqn:hinge0}. Let $\lambda' >0$ be such that $\wl = \w^*$ for all $\lambda \le \lambda'$ and $L = 2 \frac{\sum_{j=1}^n \norm{\x_j}}{n}$. For $\lambda,\widetilde \lambda\ge 0$ and data satisfying \Cref{assumption:linSep}, we have 
\begin{equation}\label{eqn:nextSolnDist}
\|\wl - \w_{\widetilde \lambda}^*\| \le  \frac{L}{2} \bigg|\frac{1}{\lambda} - \frac{1}{\tilde\lambda}\bigg|
\end{equation}
and 
\begin{equation}\label{eqn:trueSolnDist}
\|\wl - \w^*\| \le \frac{L \lambda}{2\left(\lambda'\right)^2}  .
\end{equation}
\end{lemma}

The final lemma bounds the initial error at each fixed level of regularization for the subgradient updates produced when minimizing $\Fls$. In particular, it specifies a bound shrinking in $s$ on the distance between the initial iterate $\barws$ and the minimizer $\ws$ of the function $\Fls$. The fact that the initial error for each regularized subproblem goes to zero is crucial for proving the convergence of \Cref{algo:GD} to the hard margin SVM solution.

\begin{lemma}
\label{lem:R_bound}
 Let $L = 2\frac{\sum_{j=1}^n \norm{\x_j}}{n}$ and $R_0 = \frac{L}{2\lambda_0 }$. For $s_0 \in \mathbb{N}$ with $s_0>2$, $p\in (0,1)$, and $r >2p$, let $\lambda_s = (s_0+s)^{-p}$ and $t_s = (s_0+s)^{r}$. Let 
\[
R_{s} = CL(s_0+s-1)^{-\alpha} \quad\mbox{ for }\quad  0\le \alpha \le \min \left(\frac{r-2p}{2(1+\epsilon_0)}, 1 - p  \right),
\]
with 
\[
C = \max\left\{4, \frac{1}{2\lambda_0 }(s_0-1)^{\alpha}\right\}\text{ and }\epsilon_0 = \frac{ \log(s_0 ) - \log(s_0 -1)}{\log(s_0)}.
\] 
Let $\eta_s = \frac{R_s}{L\sqrt{t_s}}$. 
Then for the averaged subgradient iterates $\barws$ of \Cref{algo:GD},
\[
\norm{ \barws - \ws} \le R_{s} .
\]
\end{lemma}

Based on \Cref{lem:R_bound}, for $r>2p$ and $p<1$ the radii $R_s$ shrink to 0 as $s$ increases.

\subsection{Proof of \Cref{thm:conv}.}\label{subsec:main_proof}
We now prove \Cref{thm:conv} using the above lemmas.
\begin{proof}
We use the triangle inequality to bound the error as
\begin{equation}\label{eqn:triangle_ineq}
\|\barws - \w^*\| \le \norm{ \barws - \ws} + \|\ws - \w^*\| .
\end{equation}
We then bound the terms $\norm{ \barws - \ws}$ and $ \|\ws - \w^*\| $ using the lemmas of \Cref{subsec:lemmas}.

Let $L = 2 \frac{\sum_{j=1}^n \norm{\x_j}}{n}$ and choose $s_0\in\mathbb{N}$ with $s_0>2$. Let $\lambda_s = (s_0+s)^{-p}$ and $t_s = (s_0+s)^{r}$. Let 
\[
R_{s} = CL(s_0+s-1)^{-\alpha} ,
\text{ for }
 \alpha = \min \left(\frac{r-2p}{2(1+\epsilon_0)}, 1 - p  \right) ,\]
with 
\[
C = \max\left\{4, \tfrac{1}{2}s_0^p(s_0-1)^{\alpha}\right\}\text{ and }\epsilon_0 = \frac{ \log(s_0 ) - \log(s_0 -1)}{\log(s_0)}.
\] 
Let $\eta_s = \frac{R_s}{L\sqrt{t_s}}$. By \Cref{lem:R_bound}, considering the first term in the bound of \Cref{eqn:triangle_ineq},
\[\norm{\barws - \ws} \le R_s = CL(s_0+s-1)^{-\alpha}. \]
Changing the base,
\[\norm{\barws - \ws} \le CL(s_0+s)^{-\alpha(1-\epsilon_0)}. \]
We now bound the second term of the bound in \Cref{eqn:triangle_ineq}. 
Let $\lambda' >0$ be such that $\wl = \w^*$ for all $\lambda \le \lambda'$. By \Cref{lem:gamma_bound},
\[ \norm{\ws- \w^*} \le \frac{L \lambda_s}{2(\lambda\rq{})^2} = \frac{L (s_0+s)^{-p}}{2(\lambda\rq{})^2} .\]

The total number of updates, $k$, used to calculate $\barws$ is bounded by 
\[k = \sum_{i=0}^{s-1} t_i = \sum_{i=s_0}^{s_0+s-1} i^r \le \int_{s_0+1}^{s+s_0} i^r = \frac{(s+s_0)^{r+1}}{r+1} . \]
Rearranging,
\[\left((r+1)k\right)^{\frac{1}{r+1}} \le s_0 + s.\]
Writing the bounds in terms of the total number of updates, $k$,
\[\norm{\barws - \ws} \le CL(s_0+s)^{-\alpha(1-\epsilon_0)} \le CL \left((r+1)k\right)^{\frac{-\alpha(1-\epsilon_0)}{r+1}} \]
and 
\[ \norm{\ws- \w^*} \le  \frac{L \left((r+1)k\right)^{\frac{-p}{r+1}} }{2(\lambda\rq{})^2} .\]
Combining these,
\begin{align*}\|\barws& - \w^*\| \le \norm{ \barws - \ws} + \|\ws - \w^*\| \\
&\le CL \left((r+1)k\right)^{\frac{-\alpha(1-\epsilon)}{r+1}} 
+  \frac{L \left((r+1)k\right)^{\frac{-p}{r+1}} }{2(\lambda\rq{})^2}.
\end{align*}

\end{proof}

In order to optimize the convergence rate given in \Cref{thm:conv}, we aim to choose parameters $p$ and $r$ such that 
\[
p,q = \argmax_{p,q} \min\left\{\frac{(r-2p)(1-\epsilon_0)}{2(1+\epsilon_0)}, (1 - p)(1-\epsilon_0), p \right\}.
\]
For $\epsilon_0$ small, $p=1/2$ and $r=2$ lead to a nearly optimal converge rate of 
\begin{align*}
\|\barws - \w^*\| 
&\le 4L \left((r+1)k\right)^{-\frac{(1-\epsilon)}{6(1+\epsilon)}} 
+  \frac{L \left((r+1)k\right)^{-1/6} }{2(\lambda\rq{})^2}.
\end{align*}
The choices $p=\tfrac{1}{2}$ and $r=2$ are considered in \Cref{cor:expl_conv} and an explicit convergence rate is given under these conditions.

\section{Experimental Results}
\label{sec:exper}
\begin{figure}
\centering
\includegraphics[width=2 in]{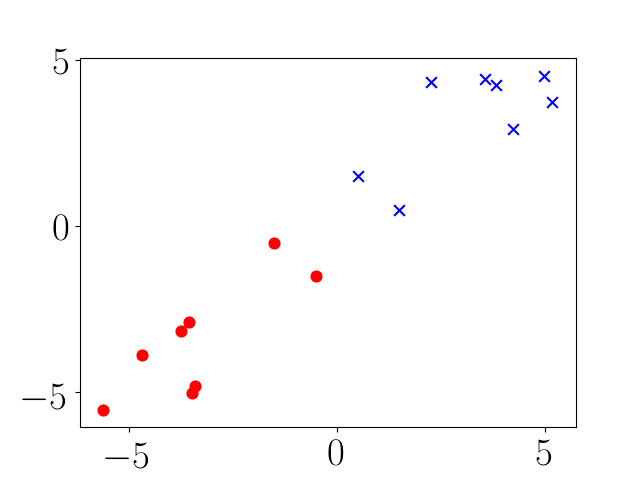}
\caption{Synthetic data considered.}
\label{fig:data}
\end{figure}

We demonstrate the convergence of \Cref{algo:GD} through several experiments on a simple synthetic dataset that is shown in \Cref{fig:data}. The experiments aim to explore the differences between convergence in theory versus practice and are not intended to be exhaustive or demonstrate superior performance over existing methods. 
The data includes four support vectors which occur at $\pm (0.5,1.5)$ and $\pm (1.5,0.5)$. The hard-margin SVM solution is given by $\w^* = (0.5,0.5)$. The maximal regularization parameter $\lambda\rq{}$ such that $\wl = \w^*$ for all $\lambda \le \lambda'$ is $\lambda' = 0.5$. We fix the parameters $p=1/2$ and $r=2$ as are considered in \Cref{cor:expl_conv} and initialize $\w_0 = \bm{0}$. 

\begin{figure}
\centering
\includegraphics[trim=0 0 0 0, clip, height = 2.2 in]{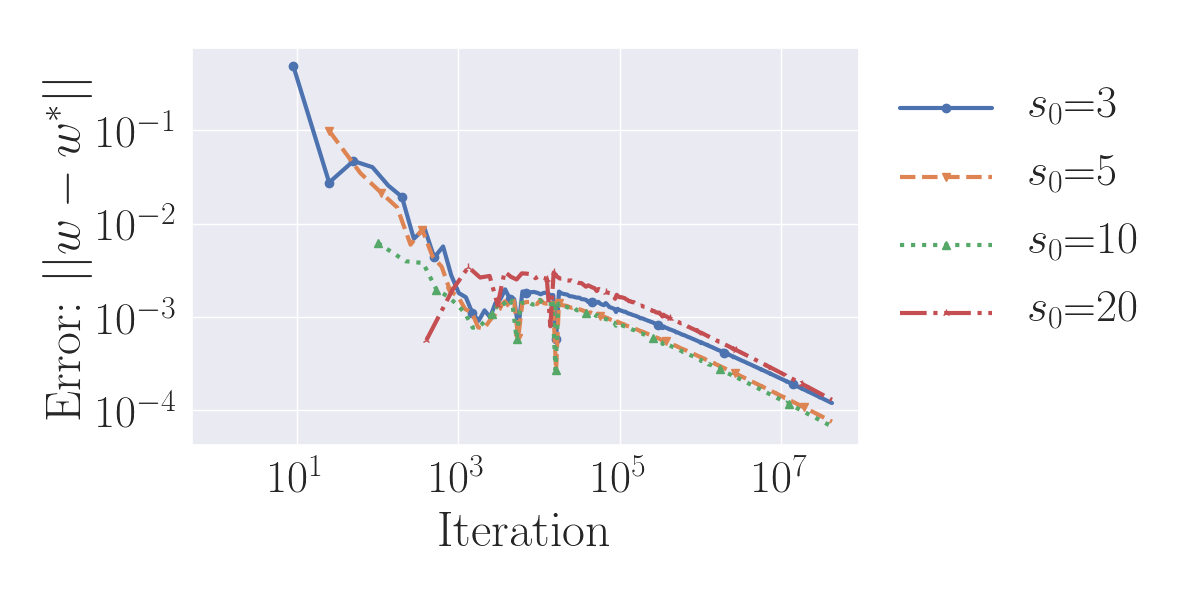}
\caption{Performance of \Cref{algo:GD} applied to data from \Cref{fig:data} with $p = 1/2$, $r = 2$ and varying $s_0$. 
}
\label{fig:1}
\end{figure}

We measure convergence in terms of the $L_2$ error as well as the angle and margin gaps of \Cref{defn:gap}. 
Convergence results for \Cref{algo:GD} with $p = 1/2$, $r=2$ and varying $s_0$ are shown in \Cref{fig:1}.  
In terms of the $L_2$ error, for a fixed number of iterations, there appears to be an optimal choice for the parameter $s_0$, as choosing $s_0= 10$ performs better than $s_0=3,5$ or $20$.

\begin{figure}
\centering
\includegraphics[trim=30 0 420 0, clip, height = 1.9 in]{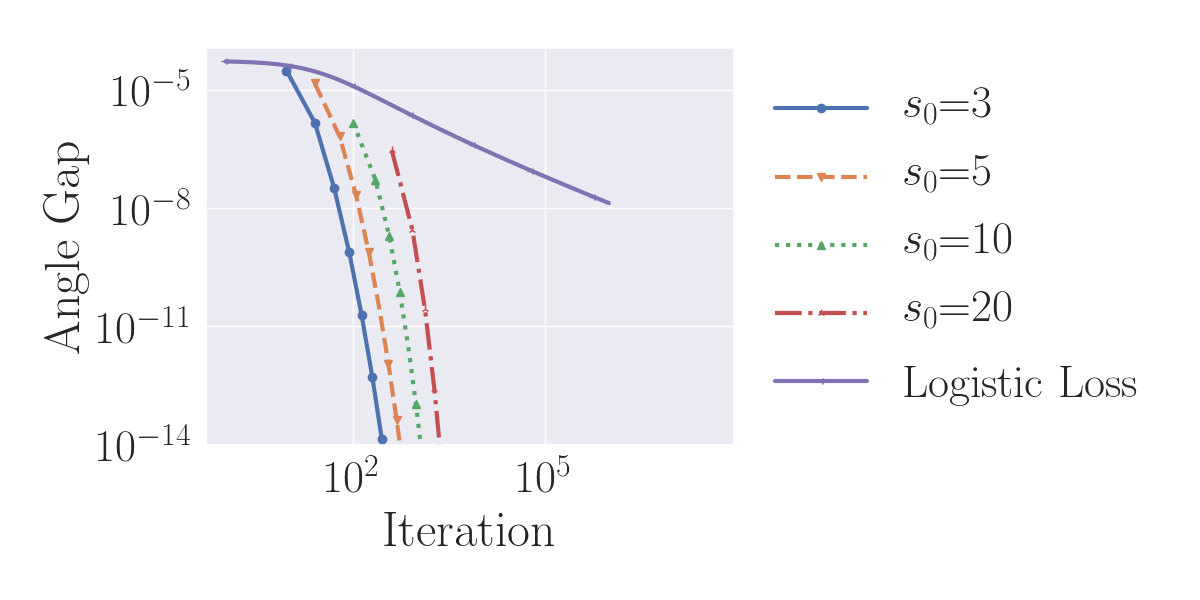}
\includegraphics[trim=30 0 420 0, clip, height = 1.9 in]{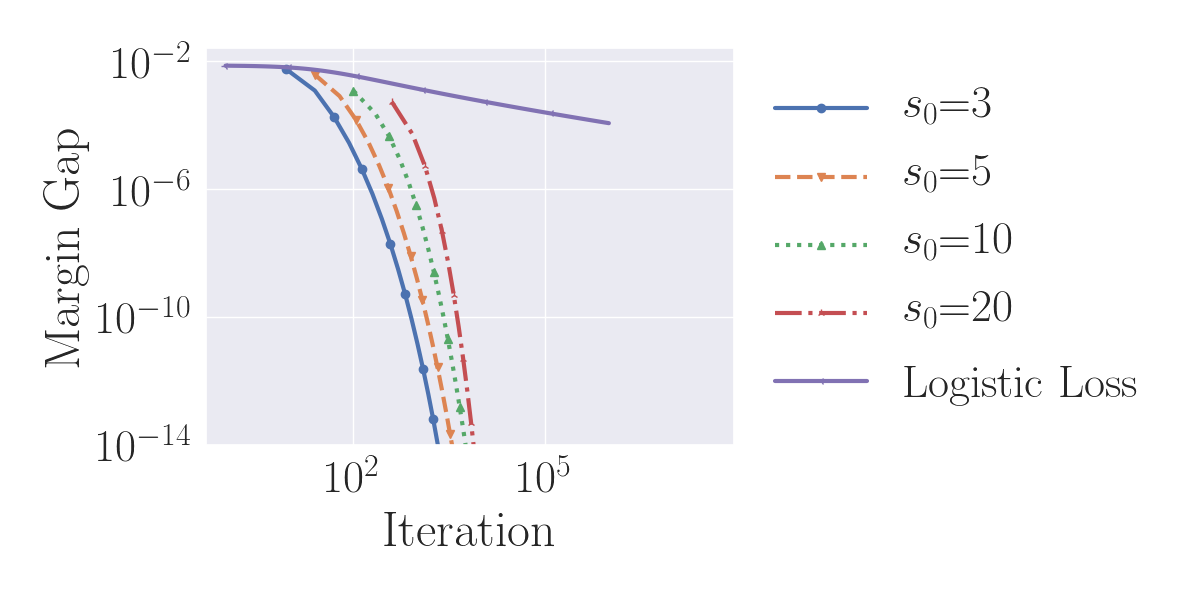}
\includegraphics[trim=550 0 45 0, clip, height = 1.6 in]{Figures/main_margin_gap.png}
\caption{Performance of \Cref{algo:GD} applied to data from \Cref{fig:data} in terms of the angle and margin gaps with $p = 1/2$, $r = 2$ and varying $s_0$. For comparison, we include gradient descent applied to the logistic loss as described in \cite{soudry2018implicit} with step size $\eta = \frac{1}{\sigma_{\max}(\X)}$, where $\sigma_{\max}(\X)$ is the largest singular value of the data matrix $\X$.}
\label{fig:1_angle_margin_gap}
\end{figure}

We additionally compare the convergence of \Cref{algo:GD} in terms of the angle gap and margin gap to gradient descent using fixed step sizes applied to the logistic loss. We use step sizes $\eta = \frac{1}{\sigma_{\max}(\X)}$, where $\sigma_{\max}(\X)$ is the largest singular value of the data matrix $\X$. As can be seen in \Cref{fig:1_angle_margin_gap}, we find significantly faster convergence via \Cref{algo:GD} as compared to minimization of the logistic loss via gradient descent with fixed step sizes as considered in \cite{soudry2018implicit,nacson2018convergence}. This result is unsurprising, as \Cref{algo:GD} arrives at the SVM solution via controlled explicit regularization as opposed to only implicit regularization via gradient descent.

\begin{figure}
\centering
\includegraphics[trim=20 0 0 0, clip, height = 2 in]{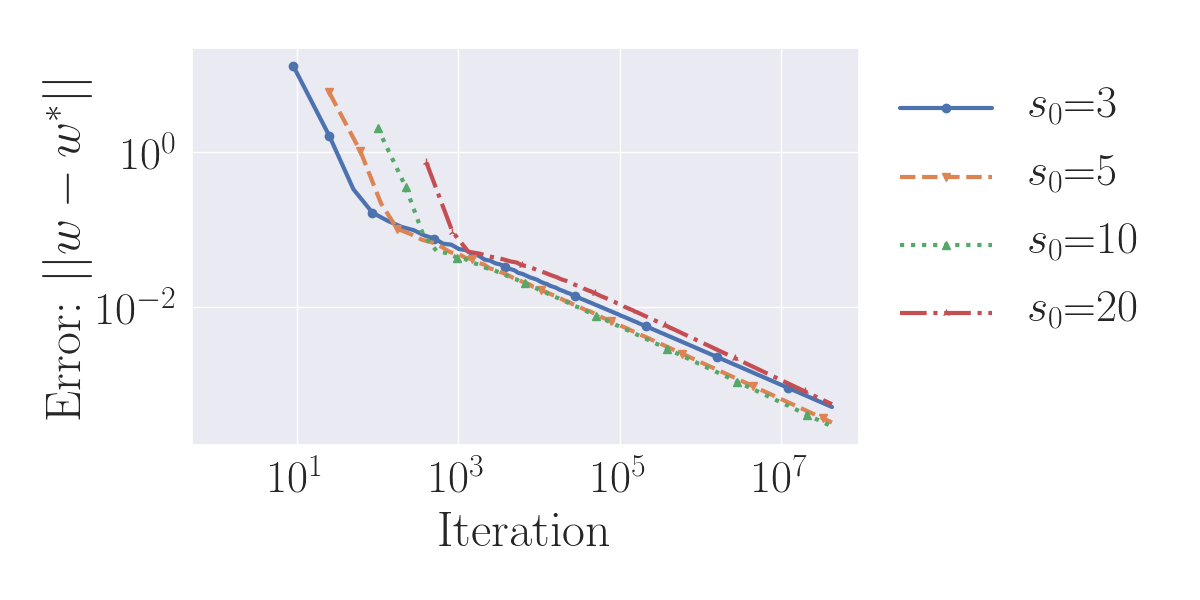}
\\
\includegraphics[trim=30 0 430 0, clip, height = 1.9 in]{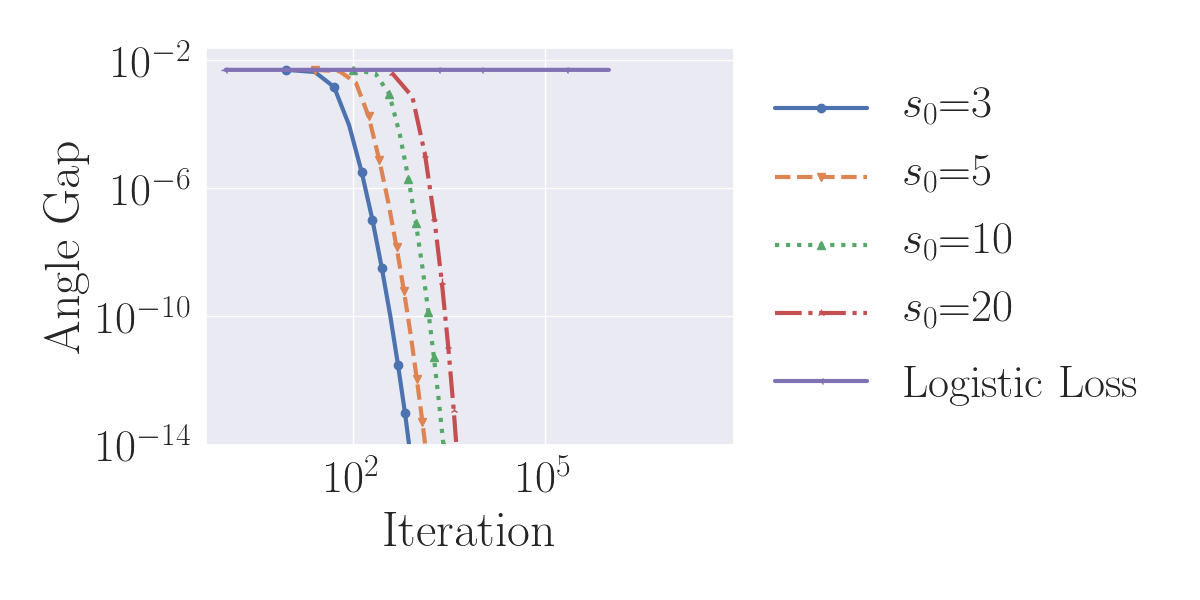}
\includegraphics[trim=30 0 430 0, clip, height = 1.9 in]{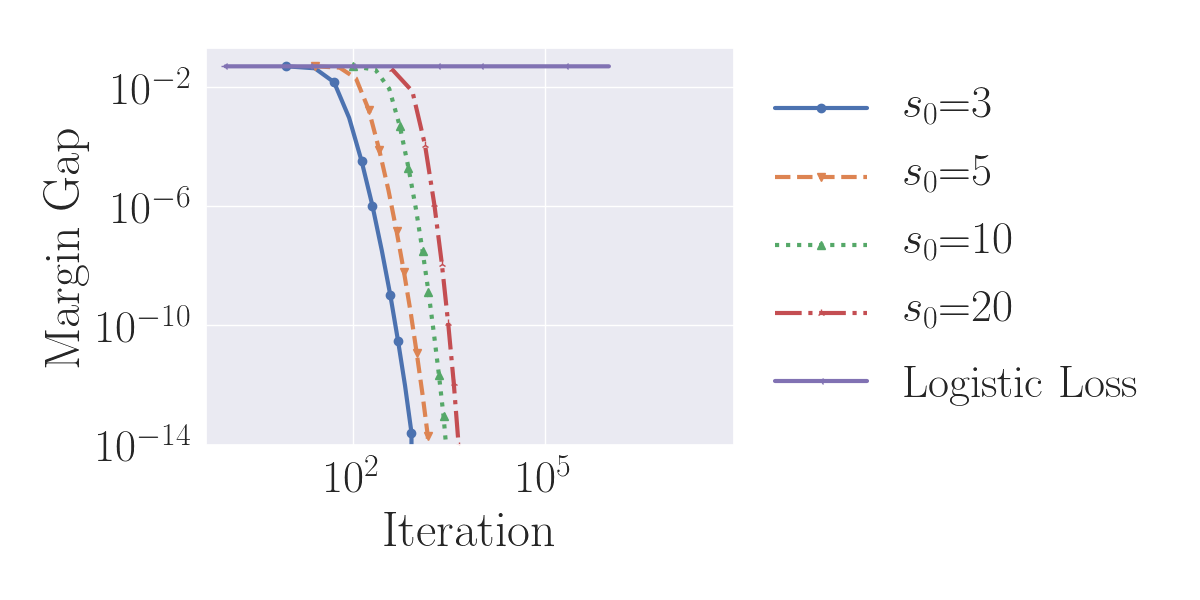}
\includegraphics[trim=570 0 45 0, clip, height = 1.7 in]{Figures/scaled_margin_gap.png}
\caption{Performance of \Cref{algo:GD} applied to the data of \Cref{fig:data}, but with the values of all y-coordinates scaled by 20. The parameters $p = 1/2$, $r = 2$ and varying $s_0$ are used. For comparison, the angle gap and margin gap plots include gradient descent applied to the logistic loss as described in \cite{soudry2018implicit} with step size $\eta = \frac{1}{\sigma_{\max}(\X)}$, where $\sigma_{\max}(\X)$ is the largest singular value of the data matrix $\X$.}
\label{fig:scaled_data}
\end{figure}

We additionally consider the performance of \Cref{algo:GD} applied to the data of \Cref{fig:data} with the y-values of the data multiplied by 20. This leads to a slightly more challenging problem with less symmetric data. The results are shown in \Cref{fig:scaled_data}. We find that the convergence of \Cref{algo:GD} is slightly slower in terms of $L_2$ error. The logistic loss converges significantly slower in terms of both the angle and margin gaps, whereas the effect on the convergence of \Cref{algo:GD} appears to be minimal.

\section{Implementation remarks}\label{sec:implement}
As presented, \Cref{algo:GD} is highly adaptable for different loss functions and settings in which one would like to consider a range of regularization parameters or variable regularization. In this section, we present several potential modifications of interest, including adaptive or gradient based step sizes, amenability to using stochastic subgradients, and alternative updates.

\subsection{Adaptive step sizes} \label{subsec:adaStep}
When the regularization parameter, $\lambda$, or the norm of $\w$ are small and close to optimal, if an iterate violates one of the hinge loss constraints, this can increase the magnitude of the gradient of the loss $\Fl$ significantly, leading to a relatively large jump in the next iterate followed by many smaller steps back toward the optimal solution of smaller norm. 
Using gradient descent with adaptive or loss-dependent step sizes can minimize the effects of these cycles. For example, we could adjust \Cref{algo:GD} to use step sizes that are normalized by the magnitude of the subgradient,
\begin{equation}
\label{eqn:aggrSteps}
\w_{k+1} = \w_k - \eta_k\frac{\nabla\Fl(\w_k)}{\|\nabla\Fl(\w_k)\|}.
\end{equation}
With this choice, the magnitude of the update is always $\eta_k$ and is independent of the magnitude of the gradient of $\Fl$. 
Cursory experimental results suggest that using adaptive step sizes as in \Cref{eqn:aggrSteps}, leads to slower convergence to the true solution initially and does not lead to improved convergence overall. 

One could also potentially increase the convergence rate guarantees for \Cref{algo:GD} by incorporating aggressive loss-dependent step sizes. In \cite{nacson2018convergence}, the authors show that when using \Cref{eqn:aggrSteps} with step sizes $\eta_k = \frac{1}{L(\w_k)}$, gradient descent applied to the logistic loss converges at the nearly optimal rate of $O(t^{-1/2}\log t)$. While this strategy provides a faster convergence rate, loss-dependent step sizes are less commonly used in practice as, in the stochastic setting, updating the loss at each iteration is often too expensive. The stochastic setting is discussed further in \Cref{subsec:stoch}.

\subsection{Regularization decay rate}
In \Cref{algo:GD}, we consider regularization parameters that decay at a rate of $\lambda_s = O(s^{-p})$ for a constant $p>0$. 
One might consider other choices for the decay rate of the regularization parameter $\lambda$. For example $\lambda_s = O\left(\tfrac{1}{\log(s)}\right)$ or $\lambda_s = O(c^s)$ for $c\in(0,1)$. 
Recall that in bounding the error $\|\barws - \w^*\|$ we use the decomposition
\[
\|\barws - \w^*\| \le \norm{ \barws - \ws} + \|\ws - \w^*\|.
\]
The first term converges more quickly when $\lambda$ is large while the second term converges more quickly when $\lambda$ is small. 
The decay rate of $\lambda_s = O(s^{-p})$ was chosen to balance the convergence of these terms.

\subsection{Stochastic subgradients}\label{subsec:stoch}
 \Cref{algo:GD} can be naturally extended to the stochastic subgradient setting, in which one performs updates based on the subgradient of the loss with respect to only a subset of the data points. This is often necessary for large-scale optimization problems. Additionally, although piecewise-constant decaying step sizes are incorporated into \Cref{algo:GD} to account for the introduced regularization, it is also often used in stochastic gradient descent in order to mitigate the effect of noise in the gradient approximation of each update \cite{bottou2018optimization}. This commonality suggests that \Cref{algo:GD} may be particularly suited for the stochastic setting.

\subsection{Alternative updates}
\begin{figure}
\centering
\includegraphics[trim=20 20 380 0, clip, height = 1.8 in]{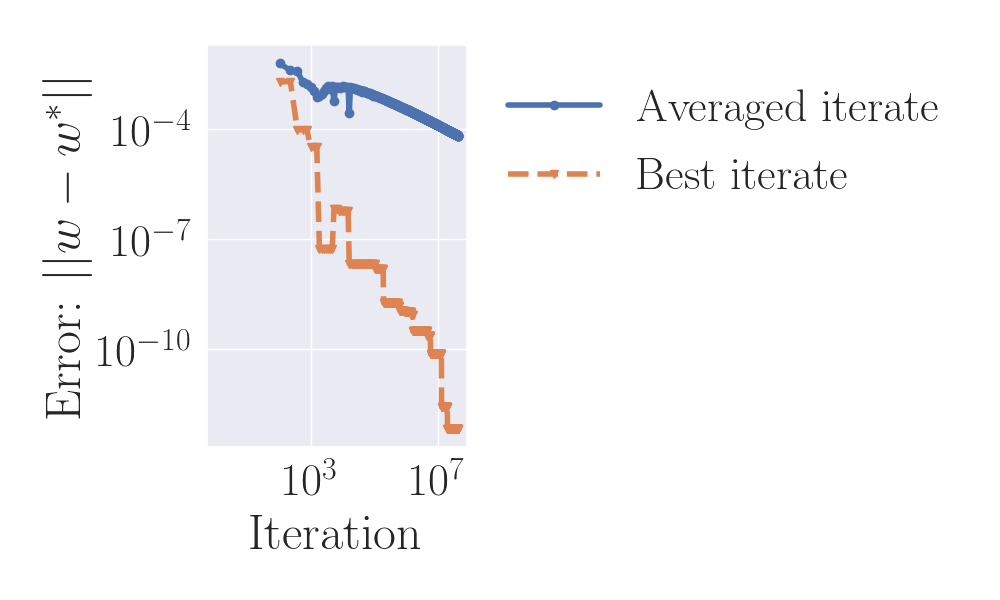}
\includegraphics[trim=20 20 380 0, clip, height = 1.8 in]{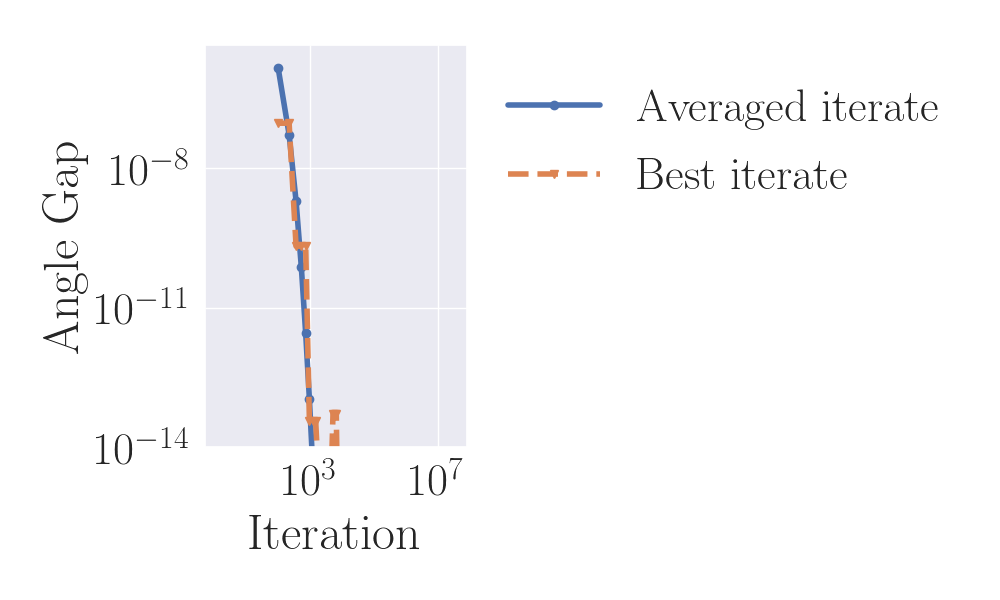}
\includegraphics[trim=20 20 380 0, clip, height = 1.8 in]{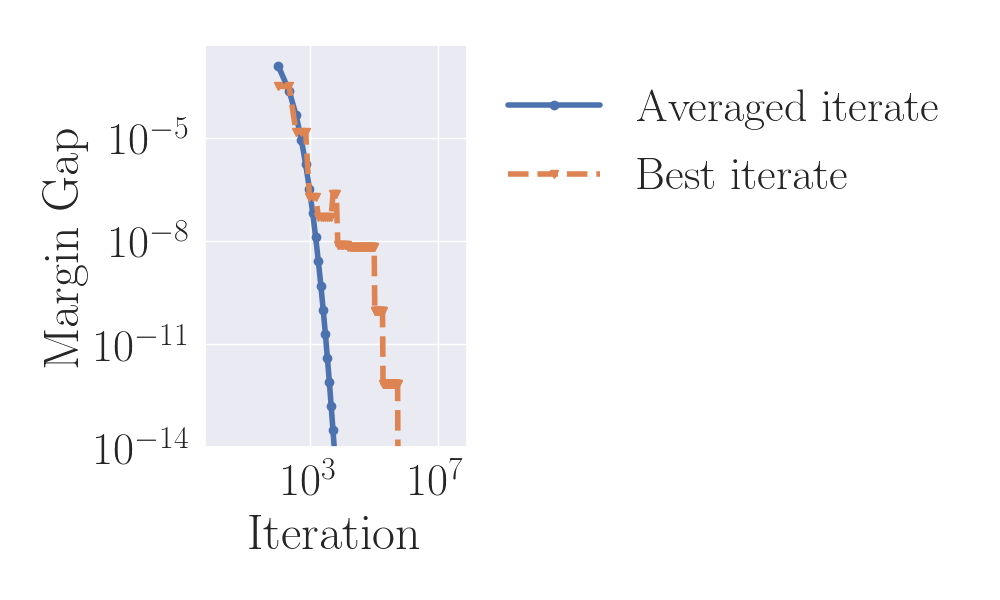}\\
\includegraphics[trim=360 270 0 40, clip, height = 0.55 in]{Figures/best_margin_gap.png}
\caption{Performance of \Cref{algo:GD} applied to data from \Cref{fig:data} using the averaged iterate versus the best iterate from each regularized subproblem. parameters $p = 1/2$, $r = 2$ and $s_0=10$ are used.  }
\label{fig:best}
\end{figure}

\Cref{lem:avg_bound} is the only result that depends on the update given by the fixed-$\lambda$ subproblem and, in particular, \Cref{thm:conv} applies to any update that satisfies $\norm{\barws - \wl} \le R_s$ for each $s = 1,\ldots, S$. Thus, as opposed to using the average of the iterates from each fixed $\lambda$ subproblem, one could use alternative updates, such as
\[\widehat \w_s = \argmin_{i=1,\dots, t_s} \Fls (\w_{i}),\]
or the iterate that leads to the minimal loss for that subproblem. We refer to this update choice as the best-iterate update and investigate the effects of this choice in \Cref{fig:best}.

 We find that the best-iterate update typically leads to significantly faster convergence in terms of the $L_2$ error. Specifically, choosing the best iterate can alleviate the slow convergence caused by the slow decrease in step size. The convergence of the two strategies, using the averaged iterate and the best iterate, perform comparably in terms of the angle gap. Using the best iterate converges somewhat slower in terms of the margin gap.

\subsection{Incorporating a bias term}
As in \cite{rosset2004margin,soudry2018implicit}, we consider the case in which the maximal-margin separating hyperplane intersects the origin. One can allow for more general hyperplanes by learning a bias term $b$ for the separating hyperplane. We propose the following method for approximating the bias term $b$
\begin{equation}\label{eqn:b_update}
b = -\frac{(\min_{i:y_i=1} \x_i^\top \w + \max_{i:y_i=-1} \x_i^\top \w)}{2},
\end{equation}
 which is guaranteed to be close to the true max-margin bias $b^*$ when $\norm{\w-\w^*}$ is small. Specifically, one can verify that for the bias $b$ as calculated in \Cref{eqn:b_update} and $b^*$ the true bias, we have 
\[|b-b^*|\le \max_i \norm{\x_i} \norm{\w-\w^*}.\]
Initial experiments with a non-trivial bias demonstrate convergence similar to the zero-bias case.

\section{Conclusion}
We have shown that, for linearly separable data, the subgradient method converges to the max-margin SVM solution when minimizing the unconstrained regularized SVM, \Cref{eqn:LShinge}, with decreasing regularization parameters, $\lambda$. Under the conditions given in \Cref{thm:conv}, this convergence can be guaranteed to be $O\left(k^{-1/6 + \delta}\right)$ for any $\delta >0$. We compare convergence rates in several metrics to those provided in \cite{soudry2018implicit,nacson2018convergence}. In particular, the convergence rate guarantees for \Cref{algo:GD} are faster than those of \cite{soudry2018implicit,nacson2018convergence} for gradient descent with fixed step sizes. This restriction to fixed or piecewise constant step sizes is a practical choice, especial when working with large-scale optimization problems. We additionally demonstrate the convergence of \Cref{algo:GD} on a simple synthetic dataset.

Although we specifically consider the hinge loss and SVMs, the results and analysis presented here could be extended to more general settings. For example, one could more generally consider settings in which one aims to solve
\[\w^* = \lim_{\lambda\to 0^+} \argmin_{\w} \frac{\lambda}{2} g(\w) + f(\w),\]
where $g$ is strongly convex and Lipschitz over bounded domains, $f$ is convex and Lipschitz, and the regularization path, 
\[\wl = \frac{\lambda}{2} g(\w) + f(\w),\] 
is Lipschitz in $\lambda$.

\acks{D.\ Molitor and D.\ Needell are grateful to and were partially supported by NSF CAREER DMS \#1348721 and
NSF BIGDATA DMS \#1740325. R.\ Ward was supported in part by AFOSR MURI Award N00014-17-S-F006.}

\bibliographystyle{myalpha}
\bibliography{bib}

\appendix

\section{Lemma Proofs}\label{sec:lemma_proofs}
We now present proofs for the lemmas of \Cref{sec:setup,sec:Main,sec:MainProof}.

We first prove \Cref{lem:iter_bound}, which gives a bound on the norm of the iterates produced by the subgradient method applied to \Cref{eqn:LShinge}.

\textbf{Proof of \Cref{lem:iter_bound}}
\begin{proof} 
Consider the subgradient update for minimizing the function $\Fl$ of \Cref{eqn:LShinge} 
\begin{equation}\label{eqn:subgrad_up_proof}
\w\rq{} = (1-\lambda\eta) \w + \frac{\eta}{n}\sum_{j : y_j\x_j^\top  \w \leq 1} y_j\x_j
\end{equation}
with $\eta\lambda < 1$. Suppose that the iterate $\w$ satisfies $\norm{\w} \le \frac{1}{\lambda n} \sum_{j=1}^n\norm{\x_j}$. We aim to show that $\w\rq{}$ given by the subgradient update also satisfies $\norm{\w}\le \frac{1}{\lambda n} \sum_{j=1}^n\norm{\x_j}.$ Taking the norm on both sides of \Cref{eqn:subgrad_up_proof},
\begin{align*}
\norm{\w\rq{}} &= \bigg\|(1-\eta\lambda)\w + \frac{\eta}{n}\sum_{j : y_j\x_j^\top  \w\leq 1} y_j\x_j \bigg \| \\
&\le (1-\eta\lambda)\norm{\w} + \frac{\eta}{n} \bigg\| \sum_{j : y_j\x_j^\top  \w\leq 1} y_j\x_j \bigg\|\\
&\le (1-\eta\lambda)\frac{1}{\lambda n} \sum_{j=1}^n\norm{\x_j} + \frac{\eta}{n}\sum_{j=1}^n \norm{\x_j}\\
&= \frac{1}{\lambda n} \sum_{j=1}^n\norm{\x_j} -\frac{\eta}{ n} \sum_{j=1}^n\norm{\x_j} + \frac{\eta}{n}\sum_{j=1}^n \norm{\x_j}\\
&= \frac{1}{\lambda n} \sum_{j=1}^n\norm{\x_j} .
\end{align*}
Thus the norms of all iterates of the subgradient method applied to the function $\Fl$ remain bounded by $\frac{1}{\lambda n} \sum_{j=1}^n\norm{\x_j}$ if the initial iterate has norm at most $\frac{1}{\lambda n}\sum_{j=1}^n\norm{\x_j}$. 
The norm of the minimizer $\wl$ of $\Fl$ must also satisfy the bound
$
 \norm{\wl} \le \frac{1}{\lambda n}\sum_{j } \norm{\x_j}
$ 
as $0\in \partial \Fl(\wl)$ and so
\[
\lambda \norm{\wl} \le \frac{1}{n}\bigg|\bigg|\sum_{j : y_j\x_j^\top  \wl \leq 1} y_j\x_j\bigg|\bigg|.
\]
\end{proof}

\textbf{Proof of \Cref{lem:angleMargin}}\\
\Cref{lem:angleMargin} uses \Cref{thm:conv} to derive bounds for the angle and margin gaps.
\begin{proof}
To derive a convergence rate for the angle gap, we use the decomposition
\begin{align*}
\|\w_k - \w^*\|^2 &= \|\w_k\|^2 + \|\w^*\|^2 - 2\w_k^\top \w^*\\
&=  \left(\|\w_k\| - \|\w^*\|\right)^2 +2\|\w_k\| \|\w^*\| - 2\w_k^\top \w^*.
\end{align*}
Dividing by $2\|\w_k\| \|\w^*\|$,
\begin{align*}
1 -  \frac{\w_k^\top \w^*}{\|\w_k\| \|\w^*\|} 
&= \frac{\|\w_k - \w^*\|^2 - \left(\|\w_k\| - \|\w^*\|\right)^2 }{2\|\w_k\| \|\w^*\|}\\
&\le \frac{\|\w_k - \w^*\|^2  }{2\|\w_k\| \|\w^*\|}
.
\end{align*}

Since $\|\w^*\|$ is necessarily bounded away from 0 since $y_i\x_i^\top \w^*\ge 1$ for all $i$. We can bound $\|\w_k\|$ away from 0 for $t$ large using the convergence of $\w_k$ to $\w^*$ guaranteed by \Cref{thm:conv}. 
Let 
\[
c =  \min \left(\frac{(r-2p)(1-\epsilon_0)}{2(r+1)(1+\epsilon_0)}, \frac{(1 - p)(1+\epsilon_0)}{r+1} , \frac{p}{r+1}  \right), 
\]
be the exponent in the convergence rate of $\|\w-\w^*\|$ and $p,r,$ and $\epsilon_0$ be defined as in \Cref{thm:conv}. 
Since 
\[(\|\w_k\| - \|\w^*\|)^2 \le \|\w_k - \w^*\|^2 \le Ak^{-2c} \]
for constants $A, c >0$ by \Cref{thm:conv}, then 
$\|\w_k\| \ge \|\w^*\| - Ak^{-c}.$ Thus for $k$ sufficiently large, we can bound $\|\w\|$ away from 0 and have 
\begin{equation}\label{eqn:angle_bound}
1 - \frac{\w_k^\top \w^*}{\|\w_k\| \|\w^*\|} = O\left(k^{-2c}\right).
\end{equation}

We now consider the margin bound.  Let $j = \argmin_{i=1,\ldots n} \frac{y_i \x_i^\top  \w_{k}}{\|\w_{k}\|}$. Since $y_i \x_i^\top  \w^*\ge 1$ for all $i = 1,\ldots, n$, we have that 
\begin{align*}
0&\le \frac{1}{\|\w^*\|}  - \frac{y_j \x_j^\top  \w_{k}}{\|\w_{k}\|} 
 \le \frac{y_j \x_j^\top  \w^*}{\|\w^*\|}  - \frac{y_j \x_j^\top  \w_{k}}{\|\w_{k}\|} \\
&=y_j \x_j^\top  \left(\frac{ \w^*}{\|\w^*\|}  - \frac{  \w_{k}}{\|\w_{k}\|} \right)
\le \|\x_j\| \bigg \| \frac{ \w^*}{\|\w^*\|}  - \frac{  \w_{k}}{\|\w_{k}\|}  \bigg\| . 
\end{align*}
Note that 
\[\bigg \| \frac{ \w^*}{\|\w^*\|}  - \frac{  \w_{k}}{\|\w_{k}\|}  \bigg\|^2 = 2 \left(1 - \frac{\w_k^\top \w^*}{\|\w_k\| \|\w^*\|}\right).\]
Assuming the data is finite and linearly separable, by \Cref{eqn:angle_bound} we then have 
\[\frac{1}{\|\w^*\|}  - \min_i \frac{y_i \x_i^\top  \w_{k}}{\|\w_{k}\|}  = O\left(k^{-c}\right). \]
\end{proof}

\textbf{Proof of \Cref{lem:avg_bound}}\\
 \Cref{lem:avg_bound} provides a modified convergence guarantee for the averaged subgradient method applied to the functions $\Fl$ \cite{bubeck2015convexarxiv}.
\begin{proof}
Let $\Fl$ be a strongly convex function with strong convexity parameter $\lambda$
and Lipschitz constant $L$ on the bounded domain considered. Let $\w_0$ be an initial iterate and $\wl$ be the minimizer of $\Fl$. Suppose $\|\w_0 - \wl \| \le R$, so that $\wl$ is contained in a ball of radius $R$ and center $\w_0$. Let $\barw = \frac{1}{t} \sum_{i=1}^t \w_i$ be the average of $t$ subgradient descent iterates with initial iterate $\w_0$ and step size $\eta = \frac{R}{L\sqrt{t}}$. We aim to show that 
\[0\le F_{\lambda} (\barw) - F_{\lambda} (\w_{\lambda}^*) \le \frac{R L}{\sqrt{t}} - \frac{\lambda}{2}\| \barw - \wl\|^2.\]
The following proof relies heavily on Theorem 3.2 of \cite{bubeck2015convexarxiv} (See also \cite{bubeck2015convex}). 

Since $\wl$ is the minimizer of $\Fl$, the inequality 
\[F_{\lambda} (\barw) - F_{\lambda} (\w_{\lambda}^*)\ge 0 \] is immediate. 
Let $g(\w) = \Fl(\w) - \frac{\lambda}{2}||\w||^2$. 
Since $g(\w)$ is convex,
\[ 
g(\barw)  \le \frac{1}{t}\sum_{i=1}^t g(\w_{i}) 
\]
and thus
\begin{align*}
\Fl & (\barw) - \frac{\lambda}{2}||\barw||^2 \le \frac{1}{t} \sum_{i=1}^t \left( \Fl(\w_{i})  -\frac{\lambda}{2}||\w_{i}||^2\right).
\end{align*}
Reorganizing and subtracting $\Fl(\wl)$,  
\begin{align}
\Fl&(\barw)-\Fl(\wl)  \nonumber\\
&\le \frac{1}{t} \sum_{i=1}^t \bigg( \Fl(\w_{i}) - \Fl(\wl) -\frac{\lambda}{2}\left(||\w_{i}||^2 -||\barw||^2\right) \bigg).\label{eqn:sum_strong_ineq}
\end{align}
Using the strong convexity of $\Fl$ and the proof of Theorem 3.2 of \cite{bubeck2015convexarxiv},
\begin{align*}
\Fl&(\w_i)-\Fl(\wl) \\
&\le \partial\Fl(\w_{i})^\top (\w_{i} - \wl) - \frac{\lambda}{2}\|\w_{i} - \wl\|^2\\
&= \frac{1}{2\eta}\left(\norm{\w_i - \w^*}^2 - \norm{\w_{i+1}-\w^*}^2\right) + \frac{\eta}{2}\norm{\partial\Fl(\w_{i})}^2- \frac{\lambda}{2}\|\w_{i} - \wl\|^2\\
&\le \frac{1}{2\eta}\left(\norm{\w_i - \w^*}^2 - \norm{\w_{i+1}-\w^*}^2\right) + \frac{\eta L^2}{2}- \frac{\lambda}{2}\|\w_{i} - \wl\|^2.
\end{align*}
Making this substitution into \Cref{eqn:sum_strong_ineq},
\begin{align*}
\Fl&(\barw)-\Fl(\wl) \\
&\le \frac{1}{2 t\eta}\left(\norm{\w_1 - \w^*}^2 - \norm{\w_{t+1}-\w^*}^2\right) + \frac{\eta L^2}{2} \\
&\quad - \frac{\lambda}{2t} \sum_{i=1}^t\bigg(\|\w_{i} - \wl\|^2 + ||\w_{i}||^2-||\barw||^2 \bigg)\\
&\le \frac{R^2}{2 t\eta} + \frac{\eta L^2}{2} - \frac{\lambda}{2t} \sum_{i=1}^t\bigg(\|\w_{i} - \wl\|^2 + ||\w_{i}||^2-||\barw||^2 \bigg)\\
&\le \frac{RL}{\sqrt{t}} - \frac{\lambda}{2t} \sum_{i=1}^t\left(||\w_{i}||^2-||\barw||^2 + \|\w_{i} - \wl\|^2\right).
\end{align*}
Decomposing the sum,
\begin{align*}
\frac{1}{t}\sum_{i=1}^t& \|\w_{i} - \wl\|^2 = \frac{1}{t}\sum_{i=1}^t \left( ||\w_{i}||^2 - 2 \w_{i}^\top \wl + ||\wl||^2\right)\\
&= \frac{1}{t}\sum_{i=1}^t  \left(||\w_{i}||^2\right)   - 2 \barw^\top \wl + ||\wl||^2\\
&= \frac{1}{t}\sum_{i=1}^t \left(||\w_{i}||^2\right)  -||\barw||^2 +||\barw||^2  - 2 \barw^\top \wl + ||\wl||^2\\
&= \frac{1}{t}\sum_{i=1}^t \left( ||\w_{i}||^2 -||\barw||^2\right)  +||\barw-\wl||^2.
\end{align*}
Making this substitution,
\begin{align*}\Fl&(\barw)-\Fl(\wl) \\
&\le \frac{RL}{\sqrt{t}} - \frac{\lambda}{t} \sum\left(||\w_{i}||^2-||\barw||^2 \right) - \frac{\lambda}{2}||\barw-\wl||^2.
\end{align*}
Since $||\w||^2$ is convex, $\frac{\lambda}{t} \sum\left(||\w_{i}||^2-||\barw||^2 \right) \ge0$ and 
\begin{align*}\Fl(\barw)-\Fl(\wl) 
&\le \frac{RL}{\sqrt{t}} - \frac{\lambda}{2}||\barw-\wl||^2
\end{align*}
as desired. 
\end{proof}

\textbf{Proof of \Cref{lem:gamma_bound}}\\
We now prove \Cref{lem:gamma_bound}, which bounds the distance between minimizers of $\Fl$ for different regularization parameters $\lambda$. 
\begin{proof}
Let $\wl$ minimize $\Fl$ as given in \Cref{eqn:LShinge}. Let $\lambda' >0$ be such that $\wl = \w^*$ for all $\lambda \le \lambda'$. For $\lambda,\widetilde \lambda\ge 0$ and data satisfying \Cref{assumption:linSep}, we aim to show that
\begin{equation*}
\|\wl - \w_{\widetilde \lambda}^*\| \le  \frac{L}{2} \bigg|\frac{1}{\lambda} - \frac{1}{\tilde\lambda}\bigg|
\end{equation*}
and 
\[
\|\wl - \w_{\widetilde \lambda}^*\| \le  \frac{\max_{j}\|\x_j\|}{\left(\lambda'\right)^2} |\lambda - \widetilde \lambda|.
\]

The proof of \Cref{lem:gamma_bound} makes use of Lemma 8 of \cite{li2018well}, which is also stated below.
\begin{lemma}
(Perturbation of strongly convex functions I \cite{li2018well}). Let $f(\z)$ be a non-negative,
$\alpha^2$-strongly convex function. Let $g(\z)$ be a L-Lipschitz non-negative convex function. For
any $\beta\ge0$, let $\z[\beta]$ be the minimizer of $f(\z) + \beta g(\z)$, then we have,
\[\bigg\|\frac{d\z[\beta]}{d\beta} \bigg \| \le \frac{L}{\alpha^2}.\]
\end{lemma}
Let $f(\w)= \|\w\|^2$ and $g(\w)=\frac{1}{n} \sum_{j=1}^n \max\{0, 1 - y_j \x_j^\top \w \}$. Then $f$ is strongly convex with strong convexity parameter $2$ and $g$ is Lipschitz with a Lipschitz constant bounded by $\tfrac{1}{n}\sum_{j=1}^n\|\x_j\|$. 
Note that 
\begin{align*}
F_\lambda(\w) &= \frac{\lambda}{2} f(\w) +  g(\w) = \frac{\lambda}{2} \left[ f(\w) + \frac{2}{\lambda}g(\w) \right]\\
&= \frac{\lambda}{2} \left[ f(\w) + \beta(\lambda)g(\w) \right]
\end{align*}
for $\beta(\lambda) = \frac{2}{\lambda}$.
Applying Lemma 8 of \cite{li2018well},
\begin{align*}
\bigg\|\frac{d\w[\lambda]}{d\lambda} \bigg \|  &
= \bigg\|\frac{d\w[\lambda]}{d\beta(\lambda)} \cdot \frac{d\beta(\lambda)}{d\lambda}\bigg \| \\
&\le  \frac{1}{2n}\sum_{j=1}^n\|\x_j\|\cdot |\beta'(\lambda)| 
= \frac{\tfrac{1}{n}\sum_{j=1}^n\|\x_j\|}{\lambda^2}.
\end{align*}

Integrating, for any $\tilde \lambda \ge \hat \lambda >0$, we have  
\begin{align*}
\norm{\w_{\tilde\lambda}^* - \w_{\hat \lambda}^*} &= \bigg\|\int_{\hat\lambda}^{\tilde\lambda} \frac{d\w[\lambda]}{d\lambda} d\lambda \bigg\|\\
&
\le   \int_{\hat\lambda}^{\tilde\lambda} \bigg\|\frac{d\w[\lambda]}{d\lambda} \bigg\| d\lambda \\
&\le \int_{\hat \lambda }^{\tilde\lambda} \frac{\tfrac{1}{n}\sum_{j} \norm{\x_j}}{\lambda^2} d\lambda \\
&
 = \tfrac{1}{n}\sum_{j} \norm{\x_j} \left |\frac{1}{\tilde \lambda} - \frac{1}{\hat \lambda}\right|.
\end{align*}

As the regularization parameter $\lambda$ approaches zero, we will use the following bound. Since for all $\lambda < \lambda'$, $\w[\lambda] = \w[\lambda'] = \w^*$, then for $\lambda < \lambda'$, $\big\|\frac{d\w[\lambda]}{d\lambda} \big \| =0$. Thus
\[\bigg\|\frac{d\w[\lambda]}{d\lambda} \bigg \| \le \frac{\max_{j}\|\x_j\|}{\left(\lambda'\right)^2} \quad \forall \; \lambda> 0.\]
This gives the second bound,
\[
\| \w_{\widetilde \lambda}^*- \w_{\hat\lambda}^*\| 
\le \int_{ \hat \lambda }^{\tilde\lambda} \frac{\tfrac{1}{n}\sum_{j} \norm{\x_j}}{\lambda^2} d\lambda
\le \int_{ \hat \lambda }^{\tilde\lambda} \frac{\tfrac{1}{n}\sum_{j} \norm{\x_j}}{\lambda\rq{}^2} d\lambda
\le  \frac{\max_{j}\|\x_j\|}{\left(\lambda'\right)^2} | \widetilde \lambda -\hat \lambda|.
\]
\end{proof}

\textbf{Proof of \Cref{lem:R_bound}}
We finally prove \Cref{lem:R_bound}, which makes use of \Cref{lem:avg_bound} and \Cref{lem:gamma_bound} to bound the initial error $\norm{\barws - \wl}$ of each regularized subproblem given in \Cref{eqn:subprob}.
\begin{proof}
We aim to show $\norm{ \barws - \ws} \le R_s$ with $R_s$ defined below and proceed by induction. For $s_0 \in \mathbb{N}$ with $s_0>2$, $p\in (0,1)$, and $r >2p$, let $\lambda_s = (s_0+s)^{-p}$, $t_s = (s_0+s)^{r}$. Recall that $L = \tfrac{2}{n} \sum_{j=1}^n\norm{\x_j}$.
For some parameter $\alpha > 0$, let
\[
R_s = CL(s_0+s-1)^{-\alpha}\text{ with }C = \max\left\{4, \frac{1}{2\lambda_0 }(s_0-1)^{\alpha}\right\}.
\]
 By \Cref{lem:iter_bound}, and since $\barw_0 = \bf{0}$, we have 
$\norm{\barw_0 - \w_{\lambda_0}^*}\le \frac{L}{2\lambda_0}$. Note that $R_0 \ge \frac{L}{2\lambda_0}$ and thus the base case, 
$
\norm{\barw_0 - \w_{\lambda_0}^*}\le R_0
$
 is satisfied.

Suppose that $\norm{ \barws - \ws} \le R_s$.  By the triangle inequality,
\[
\norm{ \barws - \ws} \le \norm{ \barws - \w_{\lambda_{s-1}}^* }+\norm{\w_{\lambda_{s-1}}^*-\ws } .
\]
For $\barws$ generated as in \Cref{algo:GD}, \Cref{lem:avg_bound} along with the inductive assumption gives that 
\[
\norm{\barws - \w_{\lambda_{s-1}}^*} \le \left(\frac{2 R_{s-1} L}{\lambda_{s-1} \sqrt{t_{s-1}}}\right)^{1/2} 
=\frac{ \sqrt{2C} L (s_0+s-2)^{-\alpha/2}}{(s_0+s-1)^{r/4-p/2}}.
\]
From \Cref{eqn:nextSolnDist} of \Cref{lem:gamma_bound},
\begin{align*}
\|\w_{\lambda_{s-1}}^*-\ws \| &\le \tfrac{L}{2}\left(\tfrac{1}{\lambda_{s}}-\tfrac{1}{\lambda_{s-1}}\right)\\
&=\tfrac{L}{2}\left((s_0+s)^p - (s_0+s-1)^p \right)\\
&\le \tfrac{Lp}{2}(s_0+s-1)^{p-1}.
\end{align*}
Combining these
\begin{align*}
\norm{ \barws - \ws}  &\le \frac{\sqrt{2C}  L (s_0+s-2)^{-\alpha/2}}{(s_0+s-1)^{r/4-p/2}}  + \frac{L}{2}p(s_0+s-1)^{p-1}.
\end{align*}

Applying a change of base via $\epsilon \ge \frac{ \log(s_0 + s -1 ) - \log(s_0 + s-2)}{\log(s_0+s - 1)}$,
\begin{align*}
\norm{ \barws - \ws} 
& \le \sqrt{2C} L(s_0+s-1)^{p/2-\alpha/2(1-\epsilon)-r/4} + \frac{Lp}{2}(s_0+s-1)^{p-1}.
\end{align*}
To simplify the analysis and remove the dependence of $\epsilon$ on the iteration number $s$, we use $\epsilon_0 = \frac{ \log(s_0 ) - \log(s_0 -1)}{\log(s_0)}.$
Now, for 
\[ 0\le \alpha \le \min \left(\frac{r-2p}{2(1+\epsilon_0)}, 1 - p  \right) \]
and $p<1$, we have 
\begin{align*}
\norm{ \barws - \ws} \le L\left(\sqrt{2C} +\frac{p}{2} \right)(s_0+s-1)^{-\alpha} 
\le CL(s_0+s-1)^{-\alpha} 
= R_{s}.
\end{align*} 
Note that allowing the first term in the upper bound on $\alpha$ to increase with $s$ leads to smaller bounds $R_s$. This choice, however, complicates the analysis. 

\end{proof}

\end{document}